\documentclass[
twocolumn,
]{ceurart}

\usepackage[most]{tcolorbox}

\begin{document}


\title{Automated Planning Techniques for Elementary Proofs in Abstract Algebra}

\author[1]{Alice Petrov}
\author[1]{Christian Muise}

\address[1]{Queen's University, 99 University Ave, Kingston, ON K7L 3N6, Canada}

\begin{abstract}
    This paper explores the application of automated planning to automated theorem proving, which is a branch of automated reasoning concerned with the development of algorithms and computer programs to construct mathematical proofs.
    In particular, we investigate the use of planning to construct elementary proofs in abstract algebra, which provides a rigorous and axiomatic framework for studying algebraic structures such as groups, rings, fields, and modules.
    We implement basic implications, equalities, and rules in both deterministic and non-deterministic domains to model commutative rings and deduce elementary results about them.
    The success of this initial implementation suggests that the well-established techniques seen in automated planning are applicable to the relatively newer field of automated theorem proving.
    Likewise, automated theorem proving provides a new, challenging domain for automated planning.
\end{abstract}


\maketitle

\section{Introduction}

Automated theorem proving is a subfield of computer science and mathematical logic that focuses on the development of algorithms and computer programs to prove mathematical theorems. 
Generally, the objective is to create a machine that can take a mathematical statement or proposition and then apply a set of logical rules to deduce a proof \cite{loveland2016automated}.
Automated theorem proving is useful in a wide range of fields, including mathematics, computer science, artificial intelligence, and philosophy, with practical applications that include the verification of software and hardware systems, the analysis of security protocols, and the development of automated reasoning systems \cite{schumann2001automated}. 

On the other hand, automated planning is a subfield of artificial intelligence that focuses on the automatic generation of plans or sequences of actions that achieve specific goals or objectives. 
The goal of automated planning is to create intelligent systems that can reason about a complex environment and come up with a set of actions to achieve a particular goal or set of goals \cite{ghallab_nau_traverso_2016}.

One of the most important challenges in automated theorem proving is to develop efficient algorithms that can handle large and complex mathematical statements. 
Many different approaches have been proposed, including deductive methods, model-based methods, and search-based methods \cite{fitting2012first}.
However, these approaches have not yet been extended to automated planning.
In this work, we formulate a number of elementary results seen in abstract algebra as planning problems, and use planners in both fully observable deterministic and non-deterministic domains to generate proofs.

The remainder of the paper is organized as follows.
Section 2 states the objectives of the work. 
Section 3 introduces the definition of commutative rings and the associated axioms.
Sections 4 and 5 outline the fully observable deterministic (FOD) model and fully observable non-deterministic (FOND) model, along with examples of proofs generated by each.
Section 6 summarizes the results. 
Section 7 discusses related work, and Section 8 discusses future work.
We conclude with Section 9.

\section{Objectives}

The objective of this paper is to demonstrate the potential application of automated planning to automated theorem proving, as well as to introduce abstract algebra as a new and challenging domain for planners.
Automated planning is a well established field, and many powerful techniques have been developed to handle the large state spaces seen in planning problems.
We show that planners are able to handle the type of axiomatic, mathematical reasoning seen in abstract algebra and find plans that correspond to reasonable proofs which obey the algebraic structure of the ambient space.
More specifically, we implement axioms, equalities, and basic implications to model the core elements and relations in commutative rings and deduce elementary results about them.
Furthermore, we demonstrate that the field is ripe with challenge problems that can serve as future benchmarks for planner development.

\section{Commutative Rings}

In abstract algebra, a commutative ring with an identity is a mathematical structure consisting of a set $R$ equipped with two binary operations, usually denoted as addition $(+)$ and multiplication $(\times)$, that satisfy the following axioms:

\begin{enumerate}
    \item [i] $R$ is closed under addition, which means that for all $a, b \in R$ we have $(a + b) \in R$.
    \item [ii] $R$ is closed under multiplication, meaning for all $a, b \in R$ we have $ab \in R$.
    \item [iii] Addition is associative, meaning that for all $a, b \in R$ we have $(a + b) + c = a + (b + c)$
    \item [iv] $R$ has a zero element, denoted by $0$, such that $a + 0 = 0 + a = a$ for all $a \in R$
    \item [v] Every element $a \in R$ has an additive inverse $-a$, such that $a + (-a) = (-a) + a = 0$
    \item [vi] Multiplication is associative, meaning that for all $a, b, c \in R$ we have $(ab)c = a(bc)$
    \item [vii] $R$ is distributive over addition, meaning that for all $a, b, c \in R$ we have $a \times (b + c) = (a \times b) + (a \times c)$ and $(a + b) \times c = (a \times c) + (b \times c)$
    \item [viii] Addition is commutative, meaning that for all $a, b \in R$ we have $a + b = b + a$
    \item [ix] Multiplication is commutative, meaning that for all $a, b \in R$ we have $a \times b = b \times a$
    \item [x]$R$ has a multiplication identity, denoted by $1$, such that $a \times 1 = 1 \times a = a$ for all $a \in R$
\end{enumerate}

Examples of commutative rings with identity include the set of integers $\mathbb{Z}$, the set of rational numbers $\mathbb{Q}$, the set of real numbers $\mathbb{R}$, and the set of complex numbers $\mathbb{C}$, among others \cite{atiyah2018introduction}. 

\section{FOD Model}

A fully observable deterministic (FOD) planning model deals with the problem of finding an optimal sequence of actions that achieve a specific goal in a fully observable and deterministic environment.
In a FOD domain, the agent has complete information about the current state of the environment, the possible actions it has available, and the consequences of those actions. 
Since the environment is deterministic, the outcome of any action is known with complete certainty \cite{ghallab_nau_traverso_2016}.

In the FOD domain, we begin by modelling the ring axioms.
For example, consider the axiom [viii], which states that addition is commutative:
$$a + b = b + a \ \text{  for all  } \ a, b \in R$$

We integrate this as the following action.
\begin{tcolorbox}[colframe=white, breakable]
\begin{footnotesize} 
\begin{verbatim}
(:action commutative-addition-axiom
    :parameters (?aPLUSb ?bPLUSa ?a ?b)
    :precondition (and 
        (issum ?aPLUSb ?a ?b) 
        (issum ?bPLUSa ?b ?a))
    :effect (and (equal ?aPLUSb ?bPLUSa)))
\end{verbatim}
\end{footnotesize}
\end{tcolorbox}
Thus, given two elements that are sums of $a$ and $b$, we set them equal.

After modelling the individual axioms, we then specify ``valid'' operations which don't require a proof to use.
For example, an element is clearly equal to itself:

$$a = a \ \text{  for all  } \ a \in R$$

We implement the statement above in the following action.
\begin{tcolorbox}[colframe=white, breakable]
\begin{footnotesize} 
\begin{verbatim}
(:action set-equal-to-self
    :parameters (?a)
    :precondition ()
    :effect (and (equal ?a ?a)))
\end{verbatim}
\end{footnotesize}
\end{tcolorbox}
In total, our FOD domain includes the following 11 predicates.
\begin{tcolorbox}[colframe=white, breakable]
\begin{footnotesize} 
\begin{verbatim}
(equal ?a ?b)
(issum ?a ?b ?c)
(iszero ?z)
(isprod ?ab ?a ?b)
(isadditiveinverse ?a ?ADDINVa)
(ismultidentity ?i)
(assumenonzero ?a)
(assumezero ?a)
(undeclared ?a)
(allowzeroprod)
(contradiction)
\end{verbatim}
\end{footnotesize}
\end{tcolorbox}
We implement a total of 28 actions in the commutative ring domain.
The first 10 actions correspond to the ring axioms.
\begin{tcolorbox}[colframe=white, breakable]
\begin{footnotesize} 
\begin{verbatim}
SECTION 0: AXIOMS
addition-axiom
multiplication-axiom
associative-addition-axiom
zero-axiom
additive-inverse-axiom
commutative-addition-axiom
associative-multiplication-axiom
multiplicative-identity-axiom
distributivity-axiom-v1
distributivity-axiom-v2
\end{verbatim}
\end{footnotesize}
\end{tcolorbox}
The next 3 actions correspond to equality operations.
\begin{tcolorbox}[colframe=white, breakable]
\begin{footnotesize} 
\begin{verbatim}
SECTION 1: EQUALITY
swap-equal
set-equal-to-self
set-equal-by-transitivity
\end{verbatim}
\end{footnotesize}
\end{tcolorbox}
These are followed by 3 actions corresponding to how we handle zeros.
\begin{tcolorbox}[colframe=white, breakable]
\begin{footnotesize} 
\begin{verbatim}
SECTION 2: ZEROS
add-zero
set-zero
set-zero-prod
\end{verbatim}
\end{footnotesize}
\end{tcolorbox}
We then implement 5 actions corresponding to operations related to sums.
\begin{tcolorbox}[colframe=white, breakable]
\begin{footnotesize} 
\begin{verbatim}
SECTION 3: SUMS
set-sum
replace-sum
swap-sum
set-equal-by-sum
add-element-to-both-sides-of-equality
\end{verbatim}
\end{footnotesize}
\end{tcolorbox}
These are followed by 5 actions corresponding to equivalent operations, but with products.
\begin{tcolorbox}[colframe=white, breakable]
\begin{footnotesize} 
\begin{verbatim}
SECTION 4: PRODUCTS
set-prod
replace-prod
swap-prod
set-equal-by-prod
multipy-element-both-sides-of-equality
\end{verbatim}
\end{footnotesize}
\end{tcolorbox}
In the final section, we implement 2 actions dealing with inverses.
\begin{tcolorbox}[colframe=white, breakable]
\begin{footnotesize} 
\begin{verbatim}
SECTION 5: INVERSES
reduce-additive-inverse
factor-out-neg
\end{verbatim}
\end{footnotesize}
\end{tcolorbox}

It is important to note that in abstract algebra, many elementary operations we take for granted must, in fact, be proven.
For example, let $R$ be a commutative ring, and let $0 \in R$ be the zero element.
For an arbitrary element $a \in R$, the result $a \times 0 = 0$ is neither an axiom nor an elementary equality and therefore requires a proof.
In the following section, we use a planner in a FOD domain for this task.

\subsection{Example Proof}

As our first example, we illustrate that additive inverses are unique.
A traditional proof of this result is as follows.

\begin{tcolorbox}[colframe=white, breakable, colback=white]
Let $R$ be a ring and let $a \in R$ be arbitrary. Suppose that $0$ is the zero element in $R$.

\vskip 0.1in

Let $b_1, b_2 \in R$ such that $a + b_1 = 0$ and $a + b_2 = 0$.  \\
We show that $b_1 = b_2$.

\vskip 0.1in

\begin{small}
$b_1 = b_1 + 0$, by the existence of the additive identity. \\
$b_1 = b_1 + (a + b_2)$, since $b_2$ is the additive inverse of $a$. \\
$b_1 = (b_1 + a) + b_2$, by the associative law of addition. \\
$b_1 = 0 + b_2$, since $b_1$ is the additive inverse of $a$. \\
$b_1 = b_2$, by the existence of the additive identity. \\
\end{small}

Therefore, there is a unique additive inverse for $a$.
\end{tcolorbox}

In order to translate this into a planning problem, we must specify an initial state and a goal.
In our initial state, we first define our zero element and then assume that both $b_1$ and $b_2$ are additive inverses of $a$.

\begin{tcolorbox}[colframe=white, breakable]

\begin{footnotesize} 
\begin{verbatim}
(:init (iszero zero)
       (is-sum zero a b1)
       (is-sum zero a b2))
\end{verbatim}
\end{footnotesize}

\end{tcolorbox}

In this case, to prove that additive inverses are unique, one may show that if both $b_1$ and $b_2$ are additive inverses of $a$ then we have $b_1 = b_2$, which implies uniqueness as desired. Thus, we have the following goal.

\begin{tcolorbox}[colframe=white, breakable]

\begin{footnotesize} 
\begin{verbatim}
(:goal (and (equal b1 b2))).
\end{verbatim}
\end{footnotesize}

\end{tcolorbox}

The LAMA planner \cite{richter2010lama} is able to find the following plan; note that we use the symbol $\implies$ to translate the line into more readable notation.

\begin{footnotesize} 
\begin{lstlisting}[breaklines=true,numbers=none, basicstyle=\footnotesize, basicstyle=\ttfamily, mathescape]
(set-equal-to-self b1)
$\implies b_1 = b_1$
(set-equal-to-self b2)
$\implies b_2 = b_2$
(swap-sum zero a b2)
$\implies 0 = b_2 + a$
(add-zero b1 b1 zero)
$\implies b_1 = b_1 + 0$
(swap-sum b1 b1 zero)
$\implies b_1 = 0 + b_1$
(add-zero b2 b2 zero)
$\implies b_2 = b_2 + 0$
(associative-addition-axiom b1 b2 zero zero b2 a b1)
$\implies (b_2 + a) + b_1 = b_2 + (a + b_1)$
$\implies 0 + b_1 = b_2 + 0$
$\implies (b_1 = b_2)$
; cost = 7 (unit cost)
\end{lstlisting}
\end{footnotesize}

\subsection{Undeclared Variables}

When writing mathematical proofs, new variables will often be introduced to define new quantities and simplify calculations.
In planning, one may wish for the planner to have some flexibility in choosing which variables to use.
We found that planners were able to handle simple proofs involving the use of an undeclared variable.
However, the introduction of more undeclared variables was a challenging addition to the problem setting.
We first introduce how undeclared variables are modeled, and then illustrate some examples of their use.

\subsubsection{Construction}

Undeclared variables are often used to model existential axioms.
For example, in a commutative ring $R$ we have that for any $a, b \in R$, there is an element $a + b \in R$.
By the ring axioms, we know such an element exists, however the precise value of this element, or whether we need to use such an element in a proof, may be unknown.
Thus, one may wish to use an undeclared variable $c$ and allow the planner to set $c = a + b$ (or any other arbitrary sum) if required.
This is an essential part of mathematical reasoning.
However, it proves difficult for planners to handle.

\subsubsection{Examples}

Consider the task of proving that, for arbitrary $a \in R$, we have
$$a \times 0 = 0$$
Our initial state is as follows.
\begin{tcolorbox}[colframe=white, breakable]
\begin{footnotesize} 
\begin{verbatim}
(:init (iszero z)
       (isprod az a z)
       (isprod minaz mina z)
       (isadditiveinverse az minaz)   
       (undeclared x))
\end{verbatim}
\end{footnotesize}
\end{tcolorbox}
Note that we add an undeclared variable $x$.
Our goal is as follows.
\begin{tcolorbox}[colframe=white, breakable]
\begin{footnotesize} 
\begin{verbatim}
(:goal (and (equal az z)))
\end{verbatim}
\end{footnotesize}
\end{tcolorbox}

The LAMA planner is able to find the following solution, in which the undeclared variable $x$ is assigned a value to aid in the proof.

\begin{footnotesize} 
\begin{lstlisting}[breaklines=true,numbers=none, basicstyle=\footnotesize, basicstyle=\ttfamily, mathescape]
(addition-axiom x az az)
$\implies x = (a \times 0) + (a \times 0)$
(additive-inverse-axiom z az minaz)
$\implies 0 = (a \times 0) - (a \times 0)$
(set-equal-to-self z)
$\implies 0 = 0$
(swap-sum z az minaz)
$\implies 0 = - (a \times 0) + (a \times 0)$
(add-zero z z z)
$\implies 0 = 0 + 0$
(distributivity-axiom-v1 az x z az az a z z)
$\implies a(0 + 0) = (a \times 0) + (a \times 0)$
(set-sum az x az az)
$\implies a \times 0 = (a \times 0) + (a \times 0)$
(reduce-additive-inverse z minaz az az az)
$\implies 0 = -(a \times 0) + (a \times 0)$ 
$= -(a \times 0) + (a \times 0) + (a \times 0) = (a \times 0)$
(swap-equal z az)
$\implies (a \times 0) = 0$
; cost = 9 (unit cost)
\end{lstlisting}
\end{footnotesize}

Now, consider the statement 
$$-1 \times a = -a$$
This is equally simple to state, however, considerably more difficult to prove.
Given all the variables it requires, the LAMA planner is able to find the following fourteen-step proof.
Note that the action (set-zero-prod) is enabled in the initial state, as this operation would otherwise require a proof to be applicable.
We include only the names of actions, and once again translate the proof into more meaningful notation.

\begin{footnotesize} 
\begin{lstlisting}[breaklines=true,numbers=none, basicstyle=\footnotesize, basicstyle=\ttfamily, mathescape]
(additive-inverse-axiom)
$\implies 0 = a + (-a)$
(additive-inverse-axiom)
$\implies 0 = 1 + (-1)$
(multiplicative-identity-axiom)
$\implies a = a \times 1$
(set-equal-by-prod)
$\implies -1 \times a = -1 \times a$
(set-zero-prod)
$\implies 0 \times a = 0 \times a$
(swap-prod)
$\implies a = 1 \times a$
(set-equal-by-prod)
$\implies a = 1 \times a$
(distributivity-axiom-v2)
$\implies 0 \times a = (1 + (-1)) \times a = (1 \times a) + (-1 \times a)$
(replace-sum)
$\implies 0 = a + (-a) = (1 \times a) + (-a)$
(swap-sum zero itimesa mina)
$\implies 0 = -a + (1 \times a)$
(set-zero)
$\implies (1 \times a) + (-1 \times a) = 0$
(additive-inverse-axiom)
$\implies (1 \times a) + (-1 \times a) = (1 \times a) + (-1 \times a) = 0$
(add-zero)
$\implies (-1 \times a) = (-1 \times a) + 0$
(reduce-additive-inverse)
$\implies (-1 \times a) = (-1 \times a) + (-a) + (1 \times a) = -a$
; cost = 14 (unit cost)
\end{lstlisting}
\end{footnotesize}

However, as soon as the variable corresponding to $(1 \times a)+(-1 \times a)$ is replaced by an undeclared variable, the planner is unable to find a proof.
This represents a novel and demanding domain in which variables are dynamically assigned values to construct a feasible path to the goal.

\section{FOND Model}

A fully observable non-deterministic (FOND) planning model deals with the problem of finding an optimal sequence of actions that achieve a specific goal in a fully observable, but non-deterministic environment.
In fully observable non-deterministic planning, the agent has complete information about the current state of the environment, the available actions, and their possible effects, but the outcome of an action is not certain. 
Instead, the environment has multiple possible outcomes, and the outcome of an action is non-deterministic \cite{ghallab_nau_traverso_2016}.

In non-deterministic planning, a controller is a component that selects actions to execute in response to observations of the environment. 
The controller uses the plan generated by the planner as a guide, selecting actions from the plan based on the current belief state and the expected outcomes of those actions \cite{ghallab2004automated}.

The branching nature of the controller makes it an interesting approach to case-based proofs.
Given a statement, one may want to check its validity on a case-by-case basis, and a carefully crafted FOND model is able to do so.
Furthermore, such a model is able to find contradictions in specific cases, and thus determine which cases are valid for the problem at hand.

\subsection{Proof by Contradiction}

In the FOND model, we introduce the notion of contradictions.
Contradictions are represented by a predicate that, once added, cannot be removed.
A predicate which indicates a contradiction is useful for both analyzing cases (which cases render our goal unsolvable) and for the general construction of a proof by contradiction.

In the example below, we use the predicates ``assumezero'' and ``iszero'' for the purpose of contradictions.
In some cases, after making no prior assumptions about a given element, we may want to set it to $0$.
If, however, we assume an element to be zero or nonzero and have an effect that changes this, we have a contradiction.
Using an extra predicate captures these assumptions.
Likewise, we use the predicate ``assumenonzero'' to capture the opposite case.

\subsection{Example Proof}

In the FOND domain, we begin by modeling the same ring axioms that we did previously.
However, suppose we now wish to model an integral domain,  which is a commutative ring $R$ with an identity such that for elements $a$ and $b$, if $ab = 0$ then $a = 0$ or $b = 0$ \cite{atiyah2018introduction}.
In particular, suppose we want to show that the cancellative law holds in an integral domain, where the cancellative law states that for a nonzero element $a \in R$, if $ab = ac$ then $b = c$.
A traditional proof of this result is as follows.

\begin{tcolorbox}[colframe=white, breakable, colback=white]
Let $a, b, c \in R$ where $R$ is an integral domain. \\
Suppose $ab = ac$ and $a \neq 0$. \\
This implies $a(b - c) = 0$. \\
Since $R$ is an integral domain this implies that either $a = 0$ or $b - c = 0$. \\
Since we assumed $a \neq 0$ we must have $b - c = 0$. \\
Therefore, $b = c$.
\end{tcolorbox}

The non-determinism in this planning problem is induced by the integral domain axiom, namely the latter $a = 0$ or $b-c = 0$ statement.
We model this in the following non-deterministic action:

\begin{tcolorbox}[colframe=white, breakable]

\begin{footnotesize} 
\begin{verbatim}
(:action integraldom-axiom
    :parameters (?ab ?a ?b)
    :precondition (and 
        (isprod ?ab ?a ?b)
        (iszero ?ab))
    :effect (oneof 
        (and 
            (when 
                (assume-nonzero ?a) 
                (contradiction))
            (iszero ?a)
            (when 
                (assume-nonzero ?b) 
                (contradiction))
            (iszero ?b)) 
        (and
            (when 
                (assume-nonzero ?a) 
                (contradiction))
            (iszero ?a) 
            (when 
                (assume-zero ?b) 
                (contradiction)) 
            (not (iszero ?b)))
        (and 
            (when 
                (assume-zero ?a) 
                (contradiction))
            (not (iszero ?a)) 
            (when 
                (assume-nonzero ?b) 
                (contradiction))
            (iszero ?b))))
\end{verbatim}
\end{footnotesize}

\end{tcolorbox}

To translate this proof into a planning problem, we first define an initial state with our zero element and our variables.
We then assume that $a$ is non-zero and $ab = ac$.

Our goal is as follows:
\begin{tcolorbox}[colframe=white, breakable]
\begin{footnotesize} 
\begin{verbatim}
(:goal (and (equal b c) (not (contradiction)))) 
\end{verbatim}
\end{footnotesize}
\end{tcolorbox}
This implies that $b = c$, and we have not reached a contradiction anywhere in our proof.

Non-deterministic actions increase the complexity of an already challenging domain.
Planners are able to find solutions to small, individual proofs.
However, sets of individual cases which call for longer paths through a non deterministic domain are more testing.
For example, the PRP (\underline{P}lanner for \underline{R}elevant \underline{P}olicies) planner \cite{muise-icaps-14} is able to find the following policy corresponding to the above task using a reduced set of actions.
\begin{figure}
  \centering
  \includegraphics[width=\linewidth]{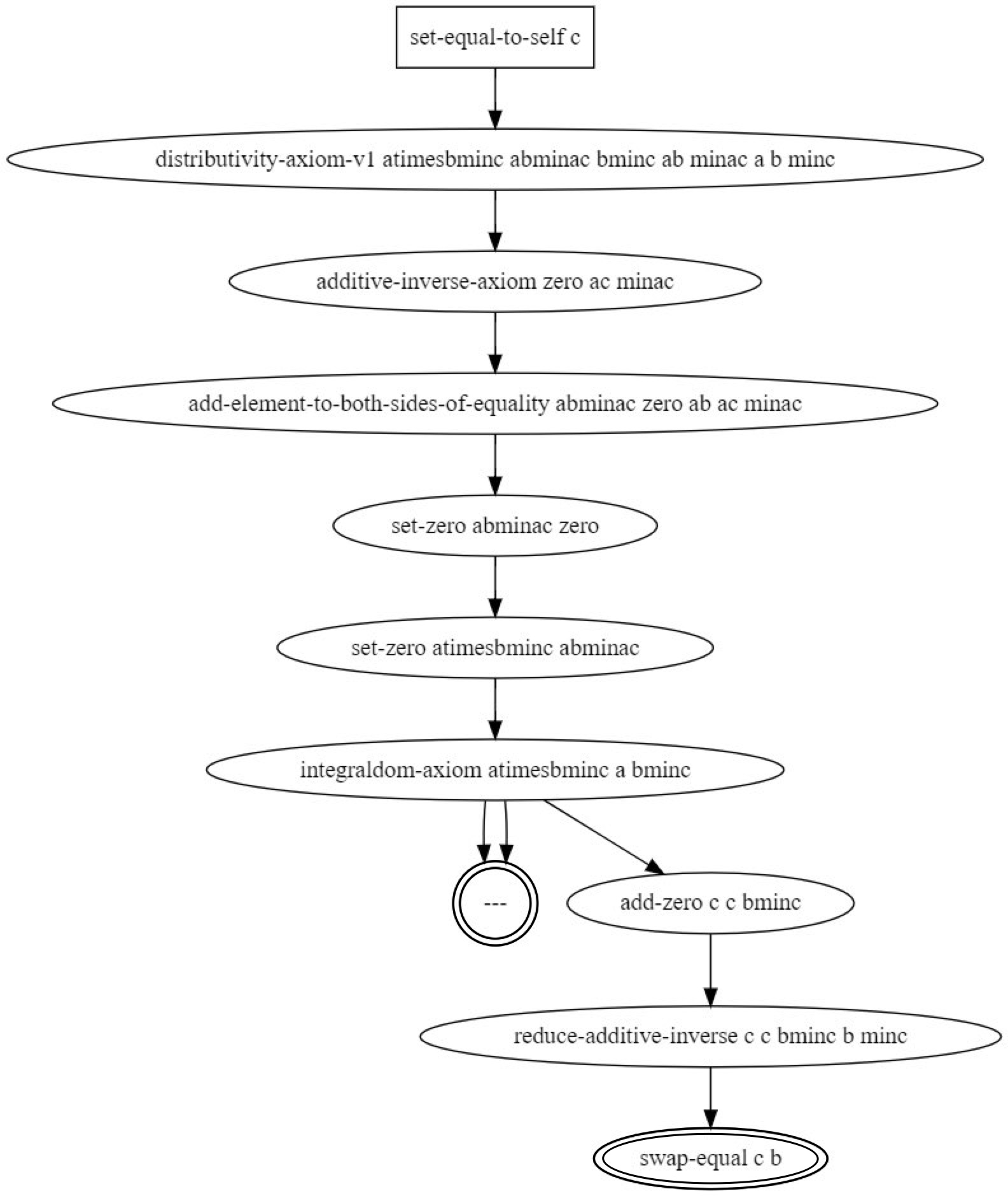}
  \caption{The policy generated by the PRP Planner corresponding to a proof that the cancellative law holds in an integral domain}
  {\label{fond-policy}}
\end{figure}
More specifically, the PRP planner constructs a policy where two out of the three possible outcomes result in a contradiction and one outcome yields the following proof.

\begin{footnotesize} 
\begin{lstlisting}[breaklines=true,numbers=none, basicstyle=\footnotesize, basicstyle=\ttfamily, mathescape]
(set-equal-to-self c)
$\implies c = c$
(distributivity-axiom-v1)
$\implies a(b - c) = ab - ac$
(additive-inverse-axiom)
$\implies 0 = ac + (-ac)$
(add-element-to-both-sides-of-equality)
$\implies ab + (-ac) = ac + (-ac)$
(set-zero)
$\implies ab + (-ac) = 0$
(set-zero)
$\implies a(b-c) = ab - ac = 0$
(integraldom-axiom)
$\implies a(b-c) = 0 \implies a = 0 \text{ or } b - c = 0$
Case 1:
$\implies a = 0 \implies contradiction \text{ (assumed } a \text{ nonzero)}$
Case 2:
$\implies b - c = 0$
(add-zero c c bminc)
$\implies c = c + 0 = c + (b - c)$
(reduce-additive-inverse c c bminc b minc)
$\implies c = b$
(swap-equal c b)
$\implies b = c$
\end{lstlisting}
\end{footnotesize}

The proof is correct, but it is not a particularly intuitive solution.
More specifically, it is not an approach a human would typically choose.
In a common approach to this proof, one would first determine that $a(b - c) = 0$, and then immediately branch off to consider the two cases $a = 0$ or $b - c = 0$.
Unlike a human, remark that the planner sets up the results it needs to use in an individual case (such as the predicate $c = c$) prior to any branching.
This occurs as there is no optimization specified to find shorter paths through the graph.
The PRP planner happened to find an ordering that worked (was allowed and achieved the goal), and so stuck with it. 
Thus, we see that a planner left to its own devices can produce correct, yet unintuitive solutions.

After the full set of actions used in the FOD domain is added back, the planner is unable to find the same solution.
Thus, we believe that longer proofs in non-deterministic domains pose an interesting challenge and are worth investigating further.

\section{Results}

In summary, we were able to model a commutative ring in both deterministic and non-deterministic domains, and deduce a number of elementary proofs based on elements and their interactions.
In the fully observable deterministic domain, we were able to prove results such as:
\begin{itemize}
    \item $0 = 0 + 0$
    \item $a \times 0 = 0$
    \item if $a + b_1 = 0$ and $a + b_2 = 0$, then $b_1 = b_2$
    \item $-1 \times a = -a$
    \item if $0 = a - b$ , then $a = b$
\end{itemize}
We were then able to extend the deterministic domain to a non-deterministic setting in order to model integral domains, and prove results such as the cancellative law, which states that if $ab = ac$ and $a \neq 0$, then $b - c$.


The notion of optimality in generated solutions is subtle and calls for some discussion.
In the domain of abstract algebra, how does one decide what a ``good'' plan is?
In classical planning, an optimal solution is typically the shortest path from the start state to the goal state.
However, writing a good proof requires readability and understandability, and it is important to keep these attributes in mind when searching for plans.
For example, when writing a simple proof one typically takes a number of ``steps'' to make a logical deduction, and then writes the desired implication as soon as the requirements to do so are met.
This makes a proof more readable, as every step is motivated.
A planner, on the other hand, has no such awareness.
For example, a planner may set a variable equal to zero far earlier than required, making it difficult for a reader to interpret the motivation for doing so (since, of course, a reader cannot see 10+ steps ahead).
A shorter plan means less ``distance'' between such logical steps and fewer items for the reader to keep track of.
Thus, readability is often a side effect of a shorter plan, but it is not strictly caused by it.
Therefore, an optimal plan may not always imply a readable proof.

However, the degree of flexibility that planning can provide has advantages.
Firstly, a planner may be able to find an unintuitive but optimal proof that a human would struggle to come up with.
With slight tweaking on the user end, such a proof can be made more readable.
Furthermore, planners can search for multiple paths in the state space, if they exist, resulting in a number of different proofs for the same problem.
If one proof is difficult to understand, it is often the case that another may be more enlightening for a reader.
Lastly, planning-based system are useful for not just generating full proofs, but suggesting potential directions to explore for finding a proof.
This can be done by relaxing the goal state and searching for partial solutions.


\section{Related Work}

Although no work has directly applied automated planning to automated theorem proving, some previous work in automated theorem proving has set out to incorporate techniques from AI.
This is particularly prevalent in proof planning, a technique for controlling search in automatic theorem proving. Proof plans capture common patterns of reasoning in a family of similar proofs and are used to guide the search for new proofs in that family. Proof plans are similar to plans constructed by plan formation techniques; however, there are differences due to the non-persistence of objects in the mathematical domain, the absence of goal interaction in mathematics, the high degree of generality of proof plans, the use of a meta-logic to describe preconditions in proof planning, and the use of annotations in formulae to guide search \cite{Bundy1996ProofP}.

In \textit{Knowledge-based proof planning} \cite{MELIS199965}, Melis and Siekmann propose a new approach to automated theorem proving that uses AI techniques such as hierarchical planning, knowledge representation, and meta-level reasoning. Unlike traditional search-based techniques, it plans the proof of a theorem at an abstract level and constructs a proof within a logical calculus using operators that represent familiar mathematical techniques. Using the OMEGA system, which implements knowledge-based proof planning, they were able to solve all well-known challenge theorems in the mathematical domain of limit theorems proposed by Woody Bledsoe \cite{melis1998heine}, including those that could not be solved by traditional systems.

Furthermore, some work has been done by Kerber et al. to integrate computer algebra in proof planning. The paper \textit{Integrating Computer Algebra into Proof Planning} \cite{Kerber1998} proposes an integration of computer algebra into mechanised reasoning systems at the proof plan level. This approach allows computer algebra algorithms to be viewed as methods, which can be used to solve problems in a certain mathematical domain. Automation is achieved by searching for a hierarchical proof plan at the method level using domain-specific control knowledge. The resulting proof plans can be expanded to produce high-level verbalized explanations or low-level machine checkable proofs.

Despite its success, there has been some critique of proof planning, summarized in \textit{A Critique of Proof Planning} \cite{Bundy2002ACO} by Dr. Alan Bundy. Proof planning aims to capture common patterns of reasoning and repair in methods and critics, which are evaluated based on criteria such as expectancy, generality, simplicity, efficiency, and parsimony. However, it is challenging to meet these criteria and avoid over-specificity or over-generalization. Overspecificity leads to a proliferation of methods and critics with limited applicability, making the resulting proof planner brittle and hard to reason about. Over-generalization, on the other hand, may permit a method to apply in an inappropriate situation. These problems can be incrementally generalized to converge on a few robust methods and critics, but this process is both infinite and non-deterministic.

Lastly, in the context of automated planning, \textit{Planning and Proof Planning} \cite{55bd780abc9548edba86065f44bcaea7} addresses the idea of knowledge-based search through proof planning. In this approach, local and global control knowledge can be expressed through a set of rules, and the design of meta-predicates that capture proof-relevant abstractions can provide an additional means of control. The article highlights the importance of hierarchical planning and the use of flexible plan patterns/macros, as demonstrated in systems like CLAM and Omega. The authors argue that experiences with proof planning systems can inform the design of planning systems that make use of similar advantages.
However, in this work we address ideas in the opposite direction and argue that experiences with planning systems can inform the search for proofs.

\section{Future Work}

This paper is intended to serve as a proof of concept, and we hope that future research will build on the ideas offered here.
To begin with, the methodology described can be extended to other algebraic structures, such as groups and fields.
Clearly, axioms will vary, but the structure of many elementary proofs in these domains remains comparable to that of commutative rings.
The introduction of a set of actions approved by the community which don't require proof would both set a standard for and aid in the development of more complex domains.

A next step in future work could involve exploring the use of different planning engines.
Some approaches may perform better than others in the presence of additional variables and undeclared variables, and it would be interesting to study how the performance of various planners is affected by the complexity of actions and the lengths of solutions.
Additionally, a number of reformulation approaches may help in making generated models easier to be dealt with by planning engines.
For example, breaking up complex actions into smaller ones which require fewer parameters may yield improved results.
Additionally, further work investigating the use of undeclared variables is required.
In the example outlined above, only one undeclared variable was used.
However, in many cases more than one may be needed, and in fact, it may not be known ahead of time how many undeclared variables are needed.
In some cases, the reformulation approaches previously mentioned may allow for better handling of undeclared variables.
This requires more investigation and scalability results.

Furthermore, one may choose to include certain elementary results in the domain without the need for a planner to prove them beforehand (i.e $a \times 0 = 0$).
These may be included as actions that implement these results, and one may use an additional predicate to toggle their applicability.
Including more operations of this kind would allow for a greater breadth and depth of proofs the planner is able to generate.
Additionally, one may model a greater number of proof methods, such as induction.
Planners begin to struggle with the addition of more complex actions (on top of the necessary axioms and elementary operations) implementing previous results.
Planners also struggled with ``forall'' statements in the goal, which in many cases are a useful way to model proofs.

Lastly, on the algebraic side, the work described here may find application in studying the underlying structure of the domain being modeled.
For example, one may investigate the longest possible plan which would correspond to a ``deep'' or non trivial result.
Multiple paths to the same goal correspond to multiple proofs of the same result, which may be of interest to study both individually and in relation to one another.
Similarly, such a tool can be used to generate homework questions for students by exploring possible paths through state spaces and the implications they induce.

\section{Conclusion}
In conclusion, this paper demonstrates the successful application of automated planning to automated theorem proving in the context of abstract algebra. 
The implementation of basic implications, equalities, and rules in both deterministic and non-deterministic domains enabled the deduction of elementary results about commutative rings. 
The results suggest that established techniques in automated planning can be effectively applied to the newer field of automated theorem proving. 
Additionally, this research highlights the potential for automated planning to tackle the challenges posed by automated theorem proving in a new way.
Conversely, abstract algebra provides a new, challenging domain for competitions and benchmarks in automated planning.
Although the LAMA and PRP planners were able to successfully find a number of elementary proofs, the introduction of undeclared variables and extended lines of reasoning in non-deterministic domains were more difficult to solve and thus provide a novel set of tasks with which to test planners.

\bibliography{refs}

\begin{thebibliography}{14}
\expandafter\ifx\csname natexlab\endcsname\relax\def\natexlab#1{#1}\fi
\providecommand{\url}[1]{\texttt{#1}}
\providecommand{\href}[2]{#2}
\providecommand{\path}[1]{#1}
\providecommand{\DOIprefix}{doi:}
\providecommand{\ArXivprefix}{arXiv:}
\providecommand{\URLprefix}{URL: }
\providecommand{\Pubmedprefix}{pmid:}
\providecommand{\doi}[1]{\href{http://dx.doi.org/#1}{\path{#1}}}
\providecommand{\Pubmed}[1]{\href{pmid:#1}{\path{#1}}}
\providecommand{\bibinfo}[2]{#2}
\ifx\xfnm\relax \def\xfnm[#1]{\unskip,\space#1}\fi
\bibitem[{Loveland(2016)}]{loveland2016automated}
\bibinfo{author}{D.~W. Loveland}, \bibinfo{title}{Automated theorem proving: A logical basis}, \bibinfo{publisher}{Elsevier}, \bibinfo{year}{2016}.
\bibitem[{Schumann(2001)}]{schumann2001automated}
\bibinfo{author}{J.~M. Schumann}, \bibinfo{title}{Automated theorem proving in software engineering}, \bibinfo{publisher}{Springer Science \& Business Media}, \bibinfo{year}{2001}.
\bibitem[{Ghallab et~al.(2016)Ghallab, Nau, and Traverso}]{ghallab_nau_traverso_2016}
\bibinfo{author}{M.~Ghallab}, \bibinfo{author}{D.~Nau}, \bibinfo{author}{P.~Traverso}, \bibinfo{title}{Automated Planning and Acting}, \bibinfo{publisher}{Cambridge University Press}, \bibinfo{year}{2016}. \DOIprefix\doi{10.1017/CBO9781139583923}.
\bibitem[{Fitting(2012)}]{fitting2012first}
\bibinfo{author}{M.~Fitting}, \bibinfo{title}{First-order logic and automated theorem proving}, \bibinfo{publisher}{Springer Science \& Business Media}, \bibinfo{year}{2012}.
\bibitem[{Atiyah(2018)}]{atiyah2018introduction}
\bibinfo{author}{M.~Atiyah}, \bibinfo{title}{Introduction to commutative algebra}, \bibinfo{publisher}{CRC Press}, \bibinfo{year}{2018}.
\bibitem[{Richter and Westphal(2010)}]{richter2010lama}
\bibinfo{author}{S.~Richter}, \bibinfo{author}{M.~Westphal},
\newblock \bibinfo{title}{The lama planner: Guiding cost-based anytime planning with landmarks},
\newblock \bibinfo{journal}{Journal of Artificial Intelligence Research} \bibinfo{volume}{39} (\bibinfo{year}{2010}) \bibinfo{pages}{127--177}.
\bibitem[{Ghallab et~al.(2004)Ghallab, Nau, and Traverso}]{ghallab2004automated}
\bibinfo{author}{M.~Ghallab}, \bibinfo{author}{D.~Nau}, \bibinfo{author}{P.~Traverso}, \bibinfo{title}{Automated Planning: theory and practice}, \bibinfo{publisher}{Elsevier}, \bibinfo{year}{2004}.
\bibitem[{Muise et~al.(2014)Muise, McIlraith, and Belle}]{muise-icaps-14}
\bibinfo{author}{C.~Muise}, \bibinfo{author}{S.~A. McIlraith}, \bibinfo{author}{V.~Belle},
\newblock \bibinfo{title}{Non-deterministic planning with conditional effects},
\newblock in: \bibinfo{booktitle}{The 24th International Conference on Automated Planning and Scheduling}, \bibinfo{year}{2014}. \URLprefix \url{http://www.haz.ca/papers/muise-icaps-14.pdf}.
\bibitem[{Bundy(1996)}]{Bundy1996ProofP}
\bibinfo{author}{A.~Bundy},
\newblock \bibinfo{title}{Proof planning},
\newblock in: \bibinfo{booktitle}{International Conference on Artificial Intelligence Planning Systems}, \bibinfo{year}{1996}.
\bibitem[{Melis and Siekmann(1999)}]{MELIS199965}
\bibinfo{author}{E.~Melis}, \bibinfo{author}{J.~Siekmann},
\newblock \bibinfo{title}{Knowledge-based proof planning},
\newblock \bibinfo{journal}{Artificial Intelligence} \bibinfo{volume}{115} (\bibinfo{year}{1999}) \bibinfo{pages}{65--105}. \URLprefix \url{https://www.sciencedirect.com/science/article/pii/S0004370299000764}. \DOIprefix\doi{https://doi.org/10.1016/S0004-3702(99)00076-4}.
\bibitem[{Melis(1998)}]{melis1998heine}
\bibinfo{author}{E.~Melis},
\newblock \bibinfo{title}{The heine--borel challenge problem. in honor of woody bledsoe},
\newblock \bibinfo{journal}{Journal of Automated Reasoning} \bibinfo{volume}{20} (\bibinfo{year}{1998}) \bibinfo{pages}{255--282}.
\bibitem[{Kerber et~al.(1998)Kerber, Kohlhase, and Sorge}]{Kerber1998}
\bibinfo{author}{M.~Kerber}, \bibinfo{author}{M.~Kohlhase}, \bibinfo{author}{V.~Sorge},
\newblock \bibinfo{title}{Integrating computer algebra into proof planning},
\newblock \bibinfo{journal}{Journal of Automated Reasoning} \bibinfo{volume}{21} (\bibinfo{year}{1998}) \bibinfo{pages}{327--355}. \URLprefix \url{https://doi.org/10.1023/A:1006059810729}. \DOIprefix\doi{10.1023/A:1006059810729}.
\bibitem[{Bundy(2002)}]{Bundy2002ACO}
\bibinfo{author}{A.~Bundy},
\newblock \bibinfo{title}{A critique of proof planning},
\newblock in: \bibinfo{booktitle}{Computational Logic: Logic Programming and Beyond}, \bibinfo{year}{2002}.
\bibitem[{Melis and Bundy(1996)}]{55bd780abc9548edba86065f44bcaea7}
\bibinfo{author}{E.~Melis}, \bibinfo{author}{A.~Bundy},
\newblock \bibinfo{title}{Planning and proof planning},
\newblock \bibinfo{year}{1996}. \bibinfo{note}{ECAI-96 Workshop on Cross-Fertilization in Planning ; Conference date: 13-08-1996}.

\end{thebibliography}

\appendix

\end{document}